\documentclass[11pt]{article}

\usepackage[margin=1in]{geometry}
\usepackage{times}
\usepackage{microtype}
\usepackage{booktabs}
\usepackage{multirow}
\usepackage{graphicx}
\usepackage{amsmath,amssymb}
\usepackage{xcolor}
\usepackage{url}
\usepackage{hyperref}
\usepackage{enumitem}
\usepackage{listings}

\hypersetup{colorlinks=true, linkcolor=blue, citecolor=blue, urlcolor=blue}
\setlist{noitemsep,topsep=2pt}

\title{SemanticZip: A Pilot Framework for Lossy Text Compression with LLMs as Semantic Decompressors}
\author{Natalia Trukhina \quad Vadim Vashkelis\\
Embedded Intelligence Lab (EMILAB)\\
\texttt{ntrukhina@emilab.org} \quad \texttt{vvashkelis@emilab.org}}
\date{May 2026}

\begin{document}
\maketitle

\begin{abstract}
Text compression for large language model (LLM) systems is usually framed as token deletion, retrieval, summarization, or exact reconstruction. We study a more aggressive but explicitly lossy setting: compress text into compact codes that an LLM can expand into task-relevant meaning. We call this setting \emph{SemanticZip}. Unlike lossless compression, SemanticZip does not require byte-identical reconstruction; unlike ordinary summarization, it treats decompression by a model as part of the codec and evaluates whether task-relevant semantic commitments are recovered.

This paper is a pilot framework, not a benchmark claim. We formalize LLM-mediated decompression, define a protected/lossy packet architecture, and evaluate six representation regimes over five author-constructed diagnostic cases: structured prose, JSON, CCL-Core, CCL-Min, SemanticZip ASCII, and SemanticZip emoji. An independent decoder LLM reconstructs typed semantic atoms from each compressed representation, and we score Critical Atom Recall (CAR), Weighted Atom Recall (WAR), precision, and tokenizer gain under \texttt{cl100k\_base} and \texttt{o200k\_base}. In this pilot, structured prose has the highest recoverability (WAR=0.956) with modest token reduction (19.1\% o200k gain). CCL-Core nearly matches recoverability (WAR=0.948) but compresses little (8.7\%). CCL-Min is the strongest balanced point (39.4\% gain, WAR=0.874). SemanticZip ASCII provides the largest useful compression (46.5\% gain) at lower recovery (WAR=0.802), while emoji-heavy SemanticZip performs worse on both compression and recovery (34.7\% gain, WAR=0.698).

The main contribution is not the claim that these numbers establish a universal frontier. They do not. Rather, the paper introduces a reproducible experimental interface for studying lossy, LLM-decompressible text codes and a design principle: safety-critical and exact commitments should remain in a protected channel, while predictable low-risk context may be semantically zipped. We identify the additional evidence required for a benchmark-level claim, including larger independently annotated data, multiple decoder models, prompt-compression baselines, and downstream task-utility validation.
\end{abstract}

\section{Introduction}

LLM applications increasingly rely on long prompts, chat histories, retrieved evidence, tool outputs, agent traces, and memory records. Context windows continue to grow, but long contexts remain costly and imperfectly used: prior work shows that model performance can degrade depending on where relevant information appears in the input \cite{liu2024lost}. Existing approaches reduce context through truncation, retrieval-augmented generation \cite{lewis2020rag}, long-context selection, natural-language summarization, memory systems \cite{packer2023memgpt}, and prompt-compression methods such as LLMLingua and LongLLMLingua \cite{jiang2023llmlingua,jiang2024longllmlingua}. These methods are useful, but many of them operate at the level of tokens, passages, or summaries rather than explicitly modeling which semantic commitments a future model response must preserve.

Our prior work, \emph{Compress the Context, Keep the Commitments} \cite{trukhina2026contextcodec}, formalized a conservative version of this problem: represent context as typed semantic atoms with canonical identity, equivalence, conflict, confidence, risk, and source spans. That work introduced Context Codec and CCL, a compact rendering of canonical JSON atoms. Its guiding question was: \emph{what must not be lost?}

This paper explores the more aggressive counterpart: \emph{how little can we say when an LLM itself is the decompressor?} We call this setting \textbf{SemanticZip}. In SemanticZip, the compressed object is not meant to reconstruct the original text exactly. Instead, it contains dense semantic cues that an LLM expands into a task-equivalent instruction, summary, or state description. For example, a travel-planning request can be compressed from prose into an ASCII code such as:

\begin{lstlisting}
@TRIP LIS/4d/Oct.early/$mod
P:walk,food.local,books,views
+Sintra BASE:Baixa|Chiado,!far
NO:nightlife,car
OUT:d2d,transit,rain,costs
\end{lstlisting}

A model-decompressor can usually infer that \texttt{LIS} means Lisbon, \texttt{4d} means four days, \texttt{\$mod} means moderate budget, and \texttt{d2d} means day-by-day output. Such compression is inherently lossy: the decompressed text may not match the source wording. The relevant question is whether it preserves the commitments needed for downstream behavior.

This framing connects to classic links between prediction and compression \cite{deletang2023compression}, approximate or semantic compression with LLMs \cite{gilbert2023semantic}, and recent prompt-compression methods \cite{jiang2023llmlingua,jiang2024longllmlingua}. Our contribution differs in focusing on explicit, human-readable, model-decompressible symbolic codes and measuring their round-trip recoverability at the level of semantic atoms.

\paragraph{Contributions.} This paper makes four contributions:
\begin{enumerate}
    \item We define SemanticZip: lossy text compression where an LLM acts as the semantic decompressor and fidelity is measured by task-relevant semantic reconstruction rather than exact text recovery.
    \item We distinguish protected commitment-preserving channels from lossy SemanticZip channels, yielding a hybrid architecture for safe context compression.
    \item We present a small experimental harness for comparing structured prose, JSON, CCL-Core, CCL-Min, SemanticZip ASCII, and SemanticZip emoji/semiotic formats through token counts and round-trip atom reconstruction.
    \item We report diagnostic results showing a coherent compression--recoverability gradient: structured representations recover best, CCL-Min provides a strong middle ground, and SemanticZip ASCII achieves the strongest useful token compression at increased semantic loss.
\end{enumerate}

\section{Background and Related Work}

\paragraph{Compression and language modeling.} The relationship between prediction and compression is classical: a strong predictive model can be converted into a compressor via coding schemes, and compression offers a lens on generalization. Deletang et al. argue that language modeling is compression and evaluate foundation models as powerful general-purpose predictors \cite{deletang2023compression}. That line of work focuses on probabilistic compression and coding. SemanticZip instead studies explicit symbolic codes that are short because an LLM can reconstruct omitted meaning from its learned priors.

\paragraph{Semantic and approximate compression.} Gilbert et al. study semantic compression with LLMs and argue that exact reconstruction is not always necessary if semantic precision or intent is preserved \cite{gilbert2023semantic}. They introduce metrics such as Exact Reconstructive Effectiveness and Semantic Reconstruction Effectiveness. SemanticZip shares the approximate-compression motivation but focuses on compressed representations designed to be decoded by a general-purpose LLM and evaluated at the level of semantic commitments.

\paragraph{Prompt compression.} LLMLingua compresses prompts through a coarse-to-fine strategy with budget control, token-level iterative compression, and distribution alignment, reporting up to 20x compression with limited performance loss in several settings \cite{jiang2023llmlingua}. LongLLMLingua targets long-context scenarios, addressing cost, latency, performance degradation, and position bias through question-aware prompt compression \cite{jiang2024longllmlingua}. Selective Context uses self-information to identify informative lexical units for efficient context management \cite{li2023selective}. These methods typically compress natural-language prompts directly. SemanticZip instead asks whether we can create compact symbolic representations that are explicitly intended for LLM-mediated decompression.

\paragraph{Long-context behavior and memory.} Lost in the Middle shows that LLMs do not always robustly use long contexts, with performance often strongest when relevant information appears near the beginning or end \cite{liu2024lost}. RAG systems ground generation in retrieved evidence \cite{lewis2020rag}, while memory-oriented systems such as MemGPT manage information across longer interactions \cite{packer2023memgpt}. SemanticZip is complementary: it can be viewed as a lossy representation layer for context that is predictable and low-risk, while exact evidence and safety-critical constraints remain protected.

\paragraph{Context Codec.} Our prior work introduced Context Codec, a formal framework for verifiable LLM context compression \cite{trukhina2026contextcodec}. It models prompts and histories as typed semantic atoms and defines metrics such as Critical Atom Recall, Weighted Atom Recall, Commitment Density, and round-trip recoverability. The current work uses that atom framework as the evaluation substrate, but studies a more aggressive lossy regime.

\paragraph{What this paper adds beyond Context Codec.} Context Codec asks which commitments must not be lost and provides a protected representation for them. SemanticZip asks the complementary question: after exact commitments have been separated, how aggressively can the remaining predictable context be compressed when the decompressor is an LLM rather than a deterministic inverse codec? The novelty here is therefore not the atom schema itself. It is the model-decompressible lossy channel, the protected/lossy packet architecture, and a pilot protocol for measuring token gain against round-trip atom recovery. In short, Context Codec is a conservative memory format; SemanticZip is a lossy channel that should be used only around such protected memory.

Prompt-compression systems are important baselines for any future empirical version of SemanticZip. LLMLingua uses a coarse-to-fine compression pipeline with budget control and token-level iterative compression to preserve semantic integrity under high compression ratios \cite{jiang2023llmlingua}. LongLLMLingua extends this line to long-context scenarios, emphasizing key-information density, position bias, and latency/cost reduction \cite{jiang2024longllmlingua}. Selective Context removes low-information lexical units to let LLMs process more content within a fixed budget \cite{li2023selective}. A full SemanticZip benchmark must compare against these systems at matched token budgets; the present paper does not yet do so.

\section{SemanticZip: Problem Formulation}

Let $x$ be an original prompt, chat-history summary, instruction, or memory state. A compressor $C$ produces a compact representation $z$:
\begin{equation}
    z = C(x, B, \pi),
\end{equation}
where $B$ is a token budget and $\pi$ is an optional domain protocol or dictionary. An LLM decompressor $D_M$ reconstructs text or atoms from $z$:
\begin{equation}
    \hat{x} = D_M(z, \pi).
\end{equation}
Traditional lossless compression asks for $\hat{x}=x$. SemanticZip relaxes this requirement. We require semantic or task-equivalent preservation:
\begin{equation}
    U(T, \hat{x}) \geq U(T, x) - \epsilon,
\end{equation}
where $T$ is the downstream task and $U$ is a task utility measure. For our diagnostic experiments, we approximate this using atom reconstruction. Let $A(x)$ be the set of gold semantic atoms in the original text and $A(D_M(z))$ the atoms reconstructed by the decompressor. SemanticZip is successful when the reconstructed atoms preserve the commitments relevant to the task.

\subsection{Semantic Atoms}

Following Context Codec \cite{trukhina2026contextcodec}, we treat a semantic atom as a typed record:
\begin{equation}
    a = (\tau, s, p, v, \mu, \sigma, e, c, r),
\end{equation}
where $\tau$ is type, $s$ subject, $p$ predicate, $v$ value, $\mu$ modality, $\sigma$ scope, $e$ evidence/source span, $c$ confidence, and $r$ risk. Examples include:

\begin{lstlisting}
{type: constraint, subject: rental_car, predicate: allowed, value: false}
{type: preference, subject: activity_style, predicate: includes, value: bookstores}
{type: output, subject: itinerary, predicate: requires, value: day_by_day}
\end{lstlisting}

\subsection{Fidelity Metrics}

For a set of critical gold atoms $A_c(x)$ and decoded atoms $\hat{A}=A(D_M(z))$, we compute Critical Atom Recall (CAR):
\begin{equation}
    \mathrm{CAR} = \frac{|A_c(x) \cap \hat{A}|}{|A_c(x)|}.
\end{equation}
Weighted Atom Recall (WAR) weights atoms by criticality $w(a)$:
\begin{equation}
    \mathrm{WAR} = \frac{\sum_{a \in A_c(x)} w(a)\mathbf{1}[a \in \hat{A}]}{\sum_{a \in A_c(x)} w(a)}.
\end{equation}
We also compute precision:
\begin{equation}
    \mathrm{Precision} = \frac{|A_c(x) \cap \hat{A}|}{|\hat{A}|}.
\end{equation}
Because LLM decoders may use different surface forms, the implementation uses normalized soft matching over type, subject, predicate, value, and scope. This evaluation is diagnostic rather than definitive; independent annotation and multi-model decompression are required for stronger empirical claims.

\section{Representation Regimes}

We compare six representation regimes.

\paragraph{Structured prose.} A carefully written natural-language compression that preserves important facts explicitly. It is easy for models to decode and often highly recoverable, but is less machine-diffable and less compact.

\paragraph{JSON.} A canonical structured representation. JSON is familiar to LLMs and compatible with validation tooling, but can be verbose as a compact rendering.

\paragraph{CCL-Core.} The conservative ASCII-first Context Compression Language from Context Codec \cite{trukhina2026contextcodec}. It is intended to be typed, auditable, and compact relative to JSON, while still readable.

\paragraph{CCL-Min.} A minified CCL profile for budget-constrained settings. It abbreviates fields and values but retains enough structure for decompression.

\paragraph{SemanticZip ASCII.} A more aggressive symbolic code using short ASCII abbreviations, compact delimiters, and domain conventions. It is lossy by design.

\paragraph{SemanticZip emoji/semiotic.} A visually compact code using emojis and semiotic symbols. The motivating example for this line of work used dense symbolic and emoji notation for a JavaScript canvas task. Our experiments test whether this style is robust compared with ASCII shorthand.

\section{Hybrid Lossy/Lossless Architecture}

SemanticZip should not compress everything. Some information is too important or rare to entrust to lossy model reconstruction. We therefore distinguish protected and lossy channels:

\begin{lstlisting}
PROTECTED: safety boundaries, exact numbers, legal/medical facts,
           private-data constraints, explicit refusals, source spans
LOSSY:     predictable defaults, background, repeated instructions,
           stylistic preferences, common schemas, low-risk summaries
\end{lstlisting}

A hybrid context packet can combine the two:

\begin{lstlisting}
@SAFE{rental_car:false; allergy:peanuts,severe; budget_max:1200_EUR}
@SZIP{TRIP:LIS/4d/Oct.early/$mod P:walk,food.local,views
      +Sintra BASE:Baixa|Chiado OUT:d2d,transit,rain,costs}
\end{lstlisting}

This architecture is the practical lesson of the experiments: high-risk commitments should remain in protected structured form, while predictable context may be compressed aggressively.

\subsection{Protected vs. Lossy Decision Rule}

An atom should be placed in the protected channel if it is safety-critical, exact, rare, legally or medically relevant, privacy-related, numerically precise, a negation, a user refusal, or a source-grounded constraint whose loss would change correctness. An atom may be placed in the lossy channel if it is low-criticality, predictable from the task schema, stylistic, redundant across context, or easy to verify after decompression. Misclassification is asymmetric: moving a protected atom into the lossy channel can weaken or erase a critical constraint, while moving a lossy atom into the protected channel mainly reduces compression. A conservative implementation should therefore default ambiguous atoms to protected storage.

\section{Experimental Setup}

\subsection{Cases}

We construct a small diagnostic set of five representative cases: Lisbon travel planning, JavaScript canvas physics code generation, Python data-cleaning script generation, React SaaS dashboard specification, and a SemanticZip research-outline prompt. Each case includes an original prompt/history, a manually annotated set of semantic atoms, and six compressed representations: structured prose, JSON, CCL-Core, CCL-Min, SemanticZip ASCII, and SemanticZip emoji.

The dataset is intentionally small and author-annotated. It should be treated as a design probe, not a definitive benchmark. The authors wrote the gold atoms and the compressed representations, so the resulting numbers can illustrate behavior of the evaluation pipeline but cannot establish independent reliability. A benchmark-level study would require blind annotation, inter-annotator agreement, held-out cases, multiple decoder models, and comparison against existing prompt-compression baselines.

\subsection{Round-Trip Decompression}

For each compressed representation, we prompt an independent decoder LLM to reconstruct semantic atoms in a canonical JSON schema. The decoder is not shown the original prompt or gold atoms. The reported pilot used a single OpenAI-family decoder configuration with temperature set to zero, top-p set to 1.0, and the prompt shown in Appendix~\ref{app:decoder-prompt}. In the experiment harness, the default decoder script uses \texttt{gpt-4o-mini}; any archival submission should replace this alias with the exact model identifier available from the provider, the API date, maximum output tokens, and the raw decoder outputs. This follows NLP reproducibility norms requiring clear model descriptions, experimental setup details, code/data availability, and reported outputs \cite{dodge2019checklist,magnusson2023reproducibility}.

We then score reconstructed atoms against gold atoms using normalized fuzzy matching over atom fields. This reduces, but does not eliminate, author-scoring circularity. In particular, the schema, gold atoms, fuzzy threshold, and atom weights remain author-defined in this pilot.

\subsection{Token Counting}

We compute token gain against the original prompt/history using two OpenAI-family tokenizers: \texttt{cl100k\_base} and \texttt{o200k\_base}. Token gain is:
\begin{equation}
    \mathrm{Gain}(z) = 1 - \frac{\mathrm{tokens}(z)}{\mathrm{tokens}(x)}.
\end{equation}
Positive values indicate compression; negative values indicate expansion.

\subsection{Atom Matching and Weights}

The pilot scorer canonicalizes decoded and gold atoms by lowercasing strings, removing punctuation, applying a small alias table for common subjects and predicates, and normalizing boolean and list-like values. Gold atoms are greedily matched to decoded atoms when a weighted similarity score exceeds 0.72. In the implementation used for Table~\ref{tab:combined}, similarity is computed as 0.40 subject match, 0.20 predicate match, 0.30 value match, and 0.10 type match. WAR weights are author-assigned criticality values on a 1--5 scale and are stored in the released case file as \texttt{criticality}. This procedure is intentionally simple and reproducible, but it is not validated as a universal semantic-equivalence metric. The 0.72 threshold is a pilot parameter, not a learned optimum. Any empirical version of this work should report threshold sensitivity, for example at 0.65, 0.72, and 0.80, because threshold choice can change absolute CAR/WAR values. The present paper therefore uses the numbers in Table~\ref{tab:combined} as a protocol demonstration rather than as threshold-independent evidence.

\subsection{Artifact Transparency}

The pilot produces five classes of artifacts: (i) the original cases and compressed representations; (ii) the gold atoms and author-assigned criticality weights; (iii) decoder input prompts; (iv) raw decoder outputs; and (v) scorer outputs and aggregate tables. To make Table~\ref{tab:combined} independently auditable, a release should include all five. In particular, the raw decoder outputs, the exact decoder model identifier, and the API date are the archival record for the decompression experiment. Without these files, the table should be read as a draft pilot result rather than a fully reproducible empirical result.

The supplementary case file exposes the weight vector used by WAR through each atom's \texttt{criticality} field. We do not claim that these weights are canonical: they are author-assigned pilot weights. A larger benchmark should report inter-annotator agreement over atom labels and weights and include unweighted CAR, weighted WAR, and critical-only recall over high-criticality atoms.

\subsection{Formats Compared}

Table~\ref{tab:format-example} shows one compact travel case across regimes. Structured prose is an editorial paraphrase baseline rather than a symbolic code. JSON is a canonical structured baseline. CCL-Core is included as a protected commitment representation from Context Codec, not because it is expected to maximize compression. CCL-Min and SemanticZip variants represent progressively more aggressive lossy renderings.

\begin{table}[t]
\centering
\small
\begin{tabular}{p{0.22\linewidth}p{0.70\linewidth}}
\toprule
Format & Example fragment \\
\midrule
Structured prose & Four-day Lisbon plan in early October; moderate budget; prefer walking, local food, bookstores, viewpoints; include Sintra; avoid nightlife-heavy plans and rental car. \\
JSON & \texttt{\{"task":"travel.plan","dest":"Lisbon","days":4,...\}} \\
CCL-Core & \texttt{@CCL/1 TASK=travel.plan DEST=Lisbon DAYS=4 PREF=\{walkable,local\_food,...\}} \\
CCL-Min & \texttt{@C1 T=trip DEST=LIS D=4 WHEN=Oct.early P=walk,foodL,books,views} \\
SemanticZip ASCII & \texttt{TRIP:LIS/4d/Oct.early/\$mod P:walk+foodL+books+views +Sintra !car} \\
SemanticZip emoji & \texttt{TRIP-PT-4D LIS OctUp EUR2 P:walk+food+books+view +Sintra no-car} \\
\bottomrule
\end{tabular}
\caption{Illustrative fragments for the travel-planning case.}
\label{tab:format-example}
\end{table}

\section{Pilot Results}\label{sec:results}

Table~\ref{tab:combined} reports the main pilot result. Structured prose gives the highest round-trip recoverability in these five cases. CCL-Core nearly matches it, JSON remains highly recoverable but does not compress under the o200k tokenizer, and CCL-Min offers the strongest balanced point. SemanticZip ASCII gives the strongest useful compression, while emoji-heavy compression performs worse than ASCII on both token gain and recovery.

\begin{table}[t]
\centering
\small
\begin{tabular}{lrrrrr}
\toprule
Format & o200k gain & cl100k gain & CAR & WAR & Precision \\
\midrule
Structured prose & 19.1\% & 18.8\% & 0.961 & 0.956 & 0.967 \\
CCL-Core & 8.7\% & 8.9\% & 0.955 & 0.948 & 0.897 \\
JSON & -3.4\% & 0.6\% & 0.944 & 0.933 & 0.894 \\
CCL-Min & 39.4\% & 40.1\% & 0.878 & 0.874 & 0.933 \\
SemanticZip ASCII & \textbf{46.5\%} & \textbf{46.1\%} & 0.794 & 0.802 & \textbf{0.975} \\
SemanticZip emoji & 34.7\% & 31.8\% & 0.684 & 0.698 & 0.928 \\
\bottomrule
\end{tabular}
\caption{Round-trip decompression and tokenizer results over five diagnostic cases. Token gain is measured relative to the original prompt/history. CAR and WAR are computed after an independent LLM decoder reconstructs canonical atoms from the compressed representation.}
\label{tab:combined}
\end{table}

\begin{figure}[t]
\centering
\includegraphics[width=0.82\linewidth]{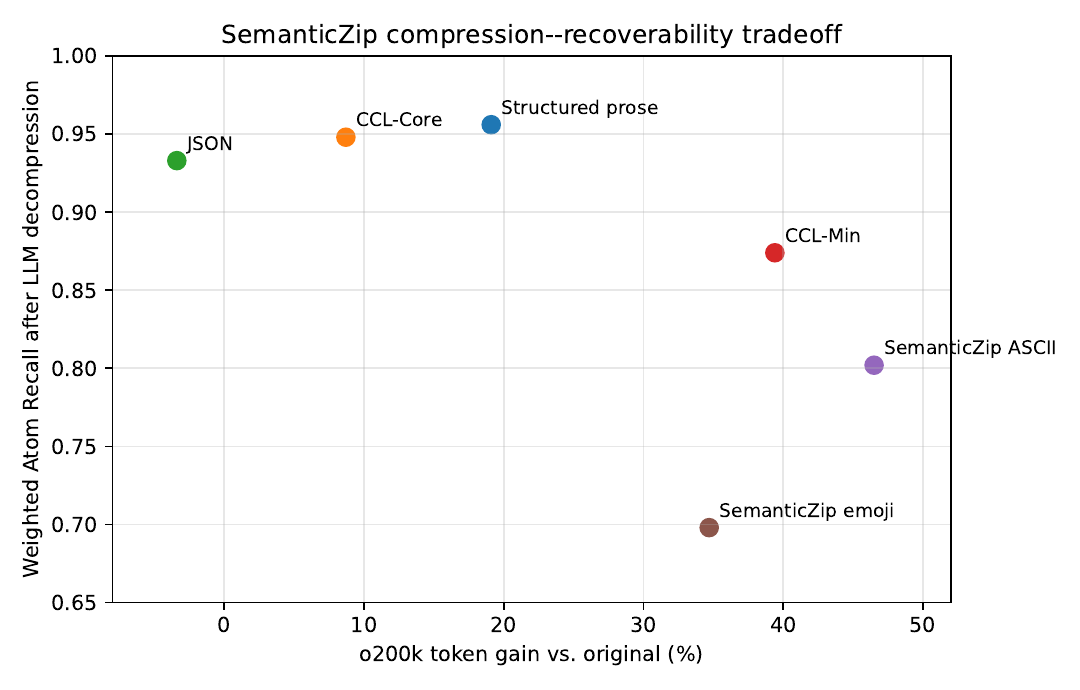}
\caption{Pilot compression--recoverability tradeoff over five author-constructed cases. SemanticZip ASCII yields the strongest token reduction among useful formats, while structured prose, CCL-Core, and JSON maximize recoverability. CCL-Min is the strongest middle point in this diagnostic run. The plot should not be interpreted as a statistically established Pareto frontier.}
\label{fig:tradeoff}
\end{figure}

\subsection{Interpretation}

The results show a pilot pattern rather than a statistically established frontier or a single universally best format. Structured prose is the recoverability ceiling in this experiment: it preserves almost all atoms and produces high precision. This is expected because natural language is the native input modality of the decoder. CCL-Core nearly matches structured prose on recall and WAR, but compresses less than expected. Its value is therefore not raw token savings; its value is typed representation, diffing, validation, conflict detection, and controlled minification.

JSON is highly recoverable but expands slightly under the o200k tokenizer. This is plausibly due to repeated field names, quotes, braces, and schema-like scaffolding; the same representation is roughly break-even under cl100k. The result is therefore tokenizer- and schema-dependent, but it supports treating JSON as a canonical internal representation rather than a compact rendering in this pilot. CCL-Min offers the best compression--recoverability middle ground, reducing tokens by roughly 39--40\% while retaining WAR=0.874. SemanticZip ASCII nearly halves token count while preserving WAR=0.802. This is too lossy for safety-critical commitments, but promising for low-risk predictable context. SemanticZip emoji is dominated by ASCII in this run: it compresses less than ASCII and recovers fewer atoms.

\subsection{Precision and Selectivity}

SemanticZip ASCII has lower recall but the highest precision. This suggests that aggressive symbolic codes tend to cause omissions rather than hallucinated commitments in this setup. That is useful for certain applications: a compressed memory may omit some preferences but is less likely to invent many unsupported facts. However, omission can still be unacceptable if the missing atom is safety-critical or numerically exact.

\section{Discussion}

\subsection{Possible Explanations for the ASCII--Emoji Gap}

The original motivation for this project included dense semiotic prompts such as:

\begin{lstlisting}
Execute - ((CODE/[rocket]/1F))
[G:m1m2/r^2; F:0.99; [lock]:PointerLock]
I:1FILE_HTML[[galaxy][comet][ball]]
...
\end{lstlisting}

Such prompts are visually compact and often surprisingly interpretable. However, the diagnostic results suggest an ASCII--emoji gap in this particular setup. This pilot does not isolate whether the weakness comes from emojis themselves, notation density, or the specific symbols selected. Two plausible explanations are as follows. First, ASCII abbreviations such as \texttt{4d}, \texttt{OUT}, \texttt{NO}, and \texttt{costs} preserve lexical anchors that are easy for tokenizers and models to process. Second, emojis can be semantically broad: a model may decode \texttt{[view]} as sunset, viewpoint, skyline, evening activity, or aesthetic preference. These are plausible explanations for higher ambiguity and lower atom recovery, but the pilot does not disentangle symbol choice from notation density or compression aggressiveness.

\subsection{When SemanticZip Makes Sense}

SemanticZip is appropriate when exact wording is not needed, the domain is familiar, the task structure is regular, and the compressed representation uses stable conventions. It is particularly promising for repeated workflows, agent memory, project-state summaries, low-risk preferences, code-generation schemas, and itinerary or planning patterns.

It is inappropriate when rare entities, legal obligations, medical facts, safety boundaries, exact numbers, or private-data constraints must be preserved. These should remain in a protected lossless or source-grounded channel. In practice, a SemanticZip system should combine aggressive lossy codes with Context Codec-style protected atoms.

\subsection{Dictionary Amortization}

A shared dictionary or protocol can improve compression after repeated use. For example:

\begin{lstlisting}
@DICT/TRIP: D=days; $mod=moderate_budget; d2d=day_by_day;
            !car=avoid_rental_car; rain=rainy_day_alternatives
\end{lstlisting}

The dictionary has an upfront token cost, but if reused across many prompts, it amortizes and improves decoding consistency. Future experiments should measure the break-even point at which a protocol dictionary outperforms ad hoc compression.

\subsection{SemanticZip and Safety}

Lossy compression is dangerous when the model's priors overwrite rare or critical details. A severe allergy, a legal exclusion, a refusal boundary, or an exact budget cap may be weakened during decompression. The system should therefore classify atoms into protected and lossy groups. Safety-relevant or high-criticality atoms should be preserved verbatim or as canonical source-grounded atoms, while SemanticZip codes should be limited to predictable low-risk content.

\section{Limitations}

This study is diagnostic and small. The cases and gold atoms were constructed by the authors, the decompression was run with a single decoder configuration, and the scorer uses normalized fuzzy matching with an author-defined alias table and threshold. The results instantiate an experimental pipeline; they do not validate that the pipeline generalizes. With five hand-crafted cases, any observed ordering may be case-dependent, decoder-dependent, or scorer-dependent. We therefore avoid statistical claims and interpret the table as a pilot pattern rather than evidence of a stable distributional frontier.

The experiments also do not include threshold sensitivity over the fuzzy matching cut-off, because the draft artifact does not embed all raw decoder outputs. This is a known reproducibility gap: if changing the cut-off from 0.72 to 0.65 or 0.80 changes the ordering of formats, the pilot should be considered threshold-sensitive. The experiments also do not measure downstream task success directly. Atom reconstruction is a proxy for utility. In some settings, a decompressed prompt may miss an atom but still produce a useful answer; in others, missing one constraint may be catastrophic. A stronger evaluation should pair atom recall with task-specific checklists or behavioral tests in the style of checklist-based NLP evaluation \cite{ribeiro2020checklist}.

The paper also lacks comparison against existing prompt-compression baselines such as LLMLingua, LongLLMLingua, and Selective Context. Therefore, the pilot should not be read as evidence that SemanticZip outperforms those methods. Its current contribution is a problem formulation, a protected/lossy architecture, and an atom-level evaluation harness.

Finally, tokenization results are model-family dependent. We report cl100k and o200k token gains, but other tokenizers may score emoji, punctuation, and abbreviations differently. Cross-model and cross-tokenizer evaluation is necessary before recommending a universal coding style.

\section{Future Work}\label{sec:future}

The next step is a larger, independently annotated benchmark with multiple decoder models, multiple tokenizers, and downstream task evaluation. Important extensions include:

\begin{enumerate}
    \item \textbf{Scale and annotation:} expand to 50--100+ cases with blind annotation, inter-annotator agreement, held-out examples, and released gold atoms.
    \item \textbf{Cross-model decompression:} test GPT-family, Claude-family, Gemini-family, Llama, Qwen, and Mistral decoders.
    \item \textbf{Learned codebooks:} induce compact domain dictionaries automatically and evaluate dictionary amortization.
    \item \textbf{Hybrid protected/lossy packets:} combine Context Codec atoms for safety-critical commitments with SemanticZip codes for predictable background context.
    \item \textbf{Baseline comparison:} compare against LLMLingua, LongLLMLingua, Selective Context, and other prompt-compression methods at matched token budgets.
    \item \textbf{Task-equivalent evaluation:} compare outputs produced from original prompts and decompressed prompts rather than atom recall alone, and report correlation between WAR/CAR and task checklist satisfaction.
    \item \textbf{Adversarial tests:} evaluate whether lossy decompression weakens negations, privacy boundaries, refusals, or exact numerical constraints.
\end{enumerate}

\section{Conclusion}

SemanticZip studies a new lossy compression regime for LLM systems: compress text into dense cues that an LLM can expand into task-equivalent meaning. In a small round-trip decompression diagnostic, structured prose and CCL-Core maximize recoverability, CCL-Min offers the strongest balanced point, and SemanticZip ASCII provides the strongest useful token compression while preserving roughly four out of five weighted atoms under this decoder and scorer. Emoji-heavy compression is less robust than ASCII shorthand in this run. The broader lesson is that context compression should be stratified: protect exact and safety-critical commitments, and use lossy model-decompressible codes only for predictable low-risk context.

\section*{Reproducibility Statement}

The experiment harness contains the five cases, compressed representations, token counting scripts, LLM decompression prompts, fuzzy atom scorer, and result aggregation scripts. Token counts were computed using cl100k and o200k tokenizers. The decoder prompts are exported as JSONL before model invocation so that any model provider can be substituted. For archival reproducibility, a repository should include the raw decoder outputs, exact model identifier, API date, maximum output tokens, scorer code, alias table, atom weights, threshold-sensitivity script, and all generated tables. The round-trip decompression table in this draft should be treated as diagnostic; exact scores may vary with decoder model, model version, prompting, temperature, maximum output length, normalization threshold, and alias table.

\appendix
\section{Decoder Prompt}\label{app:decoder-prompt}

The following prompt template was used to generate atom reconstructions from each compressed representation. The decoder was not shown the original prompt or gold atoms.

\begin{lstlisting}
Given the compressed context below, reconstruct the semantic atoms as JSON.
Return only valid JSON in this exact form:
{
  "atoms": [
    {
      "type": "constraint|goal|entity|preference|decision|procedure|output|safety",
      "subject": "canonical_snake_case_subject",
      "predicate": "equals|allowed|required|preferred|includes",
      "value": "string_or_boolean_or_number",
      "modality": "must|should|may|unknown",
      "scope": "task|output|artifact|trip|code|unknown"
    }
  ]
}
Use snake_case for subjects and values. Convert negations into predicate/value pairs.
For example, "no external libraries" becomes subject="external_libraries", predicate="allowed", value=false.
Do not infer facts not supported by the compressed context.

Compressed context:
<COMPRESSED_REPRESENTATION>
\end{lstlisting}

\section{Fuzzy Matching Details}\label{app:fuzzy}

For the pilot scorer, each atom is normalized by lowercasing strings, removing punctuation, normalizing booleans and numbers, and applying a small alias table for common fields such as \texttt{libs} $\rightarrow$ \texttt{external\_libraries} and \texttt{d2d} $\rightarrow$ \texttt{day\_by\_day}. Similarity between a gold atom and a decoded atom is:
\begin{equation}
0.40\,s_{subject} + 0.20\,s_{predicate} + 0.30\,s_{value} + 0.10\,s_{type}.
\end{equation}
Gold atoms are greedily matched to unused decoded atoms when similarity is at least 0.72. This threshold was chosen as a practical pilot setting rather than learned from validation data. Because the match decision is thresholded, an empirical submission should include a sensitivity table over at least three cut-offs (e.g., 0.65, 0.72, and 0.80) and report whether the relative ordering of formats remains stable. Threshold tuning is a modeling choice, not an invariant property of the data; this mirrors standard classifier evaluation practice where decision thresholds may be tuned for a target metric or cost \cite{sklearnthreshold}.


\begin{thebibliography}{99}

\bibitem{gilbert2023semantic}
Henry Gilbert, Michael Sandborn, Douglas C. Schmidt, Jesse Spencer-Smith, and Jules White.
\newblock Semantic compression with large language models.
\newblock arXiv preprint arXiv:2304.12512, 2023.
\newblock \url{https://arxiv.org/abs/2304.12512}.

\bibitem{deletang2023compression}
Gr{\'e}goire Del{\'e}tang, Anian Ruoss, Paul-Ambroise Duquenne, Elliot Catt, Tim Genewein, Christopher Mattern, Jordi Grau-Moya, Li Kevin Wenliang, Matthew Aitchison, Laurent Orseau, Marcus Hutter, and Joel Veness.
\newblock Language modeling is compression.
\newblock arXiv preprint arXiv:2309.10668, 2023.
\newblock \url{https://arxiv.org/abs/2309.10668}.

\bibitem{jiang2023llmlingua}
Huiqiang Jiang, Qianhui Wu, Chin-Yew Lin, Yuqing Yang, and Lili Qiu.
\newblock LLMLingua: Compressing prompts for accelerated inference of large language models.
\newblock In \emph{Proceedings of EMNLP}, 2023.
\newblock \url{https://aclanthology.org/2023.emnlp-main.825/}.

\bibitem{jiang2024longllmlingua}
Huiqiang Jiang, Qianhui Wu, Xufang Luo, Dongsheng Li, Chin-Yew Lin, Yuqing Yang, and Lili Qiu.
\newblock LongLLMLingua: Accelerating and enhancing LLMs in long context scenarios via prompt compression.
\newblock In \emph{Proceedings of ACL}, 2024.
\newblock \url{https://arxiv.org/abs/2310.06839}.

\bibitem{li2023selective}
Yucheng Li.
\newblock Selective Context: Compress your input to ChatGPT or other LLMs.
\newblock 2023.
\newblock \url{https://github.com/liyucheng09/Selective_Context}.

\bibitem{liu2024lost}
Nelson F. Liu, Kevin Lin, John Hewitt, Ashwin Paranjape, Michele Bevilacqua, Fabio Petroni, and Percy Liang.
\newblock Lost in the middle: How language models use long contexts.
\newblock \emph{Transactions of the Association for Computational Linguistics}, 12:157--173, 2024.
\newblock \url{https://aclanthology.org/2024.tacl-1.9/}.

\bibitem{lewis2020rag}
Patrick Lewis, Ethan Perez, Aleksandra Piktus, Fabio Petroni, Vladimir Karpukhin, Naman Goyal, Heinrich K{\"u}ttler, Mike Lewis, Wen-tau Yih, Tim Rockt{\"a}schel, Sebastian Riedel, and Douwe Kiela.
\newblock Retrieval-augmented generation for knowledge-intensive NLP tasks.
\newblock In \emph{Advances in Neural Information Processing Systems}, 2020.
\newblock \url{https://arxiv.org/abs/2005.11401}.

\bibitem{packer2023memgpt}
Charles Packer, Vivian Fang, Shishir G. Patil, Kevin Lin, Sarah Wooders, and Joseph E. Gonzalez.
\newblock MemGPT: Towards LLMs as operating systems.
\newblock arXiv preprint arXiv:2310.08560, 2023.
\newblock \url{https://arxiv.org/abs/2310.08560}.

\bibitem{trukhina2026contextcodec}
Natalia Trukhina and Vadim Vashkelis.
\newblock Compress the context, keep the commitments: A formal framework for verifiable LLM context compression.
\newblock arXiv preprint arXiv:2605.17304, 2026.
\newblock \url{https://arxiv.org/abs/2605.17304}.

\bibitem{jsonschema2020}
JSON Schema Organization.
\newblock JSON Schema validation: A vocabulary for structural validation of JSON.
\newblock Draft 2020-12, 2020.
\newblock \url{https://json-schema.org/draft/2020-12/json-schema-validation}.

\bibitem{rfc6902}
Paul Bryan and Mark Nottingham.
\newblock JavaScript Object Notation (JSON) Patch.
\newblock RFC 6902, Internet Engineering Task Force, 2013.
\newblock \url{https://datatracker.ietf.org/doc/html/rfc6902}.


\bibitem{sklearnthreshold}
Scikit-learn developers.
\newblock Tuning the decision threshold for class prediction.
\newblock scikit-learn User Guide.
\newblock \url{https://scikit-learn.org/stable/modules/classification_threshold.html}.

\bibitem{dodge2019checklist}
Jesse Dodge, Suchin Gururangan, Dallas Card, Roy Schwartz, and Noah A. Smith.
\newblock Show your work: Improved reporting of experimental results.
\newblock \emph{EMNLP-IJCNLP}, 2019. See also NLP Reproducibility Checklist.
\newblock \url{https://www.jessedodge.ai/NLP_Reproducibility_Checklist_V1.2.pdf}.

\bibitem{magnusson2023reproducibility}
Ian Magnusson, Akshita Bhagia, Valentin Hofmann, Luca Soldaini, Oyvind Tafjord, Peter West, Kyle Lo, Dirk Groeneveld, Kyle Richardson, Ashish Sabharwal, Iz Beltagy, and Jesse Dodge.
\newblock Reproducibility in NLP: What have we learned from the checklist?
\newblock \emph{Findings of ACL}, 2023.
\newblock \url{https://aclanthology.org/2023.findings-acl.809/}.

\bibitem{ribeiro2020checklist}
Marco Tulio Ribeiro, Tongshuang Wu, Carlos Guestrin, and Sameer Singh.
\newblock Beyond accuracy: Behavioral testing of NLP models with CheckList.
\newblock \emph{ACL}, 2020.
\newblock \url{https://aclanthology.org/2020.acl-main.442/}.
\end{thebibliography}
\end{document}